\documentclass[letterpaper, 10 pt, conference]{ieeeconf}  

\IEEEoverridecommandlockouts                              

\usepackage{datetime2}
\usepackage{cite}
\usepackage{textcomp}
\usepackage{xcolor}
\usepackage{titlesec}
\titlespacing*{\subsubsection}{0pt}{0.5em}{0.5em}
\titleformat{\subsubsection}[runin]
  {\normalfont\normalsize\bfseries}{}{0em}{}[.]

\usepackage{xcolor}
\usepackage[normalem]{ulem}
\usepackage[font=footnotesize]{caption}
\usepackage{amsmath,amssymb,amsfonts}

\usepackage{amsthm}
\usepackage{algorithm}
\usepackage{algpseudocode}
\newcommand{\NoNumberEndFor}{\let\oldalglinenumber\alglinenumber\def\alglinenumber##1{}\EndFor\let\alglinenumber\oldalglinenumber\addtocounter{ALG@line}{-1}}

\usepackage{tikz}
\usepackage{upgreek}
\usepackage{textgreek}
\usepackage{mathtools,mathdots}
\usepackage{xparse}
\usepackage{nameref}
\usepackage{graphicx}
\usepackage{float}
\usepackage{nicefrac}
\usepackage{multirow}
\usepackage{booktabs}
\usepackage{multicol}
\usepackage{hyperref}
\usepackage{multimedia}
\usepackage{arydshln}
\usepackage{substr}
\usepackage{mathrsfs}
\usepackage{pbox}
\usepackage{cleveref}
\usepackage{bm}

\let\oldmathcal=\mathcal
\renewcommand{\mathcal}[1]{
    \IfSubStringInString{#1}{ABCDEFGHIJKLMNOPQRSTUVWXYZ}{\oldmathcal{#1}}{
    \IfSubStringInString{#1}{abcdefghijklmnopqrstuvwxyz}{\mathsf{#1}}{
    \ifthenelse{\equal{#1}{\epsilon}}{\textnormal{\straightepsilon}}{
    #1
    }}}
}

\algrenewcommand\alglinenumber[1]{#1:}

\DeclareMathOperator{\arctan2}{arctan2}

\newcommand{\defined}{\vcentcolon=}

\newcommand{\reals}{\mathbb{R}}

\newcommand{\Mat}[1][]{\ifthenelse{\equal{#1}{}}{\text{Mat}}{\text{Mat}(#1)}}

\newcommand{\SE}[1]{\text{SE}(#1)}

\newcommand{\TSE}[2][]{
	\ifthenelse{\equal{#1}{}}
	{{T\SE{#2}}}
	{{T_{#1}\SE{#2}}}
}
\newcommand{\dualTSE}[2][]{
	\ifthenelse{\equal{#1}{}}
	{{T^*\SE{#2}}}
	{{T^*_{#1}\SE{#2}}}
}

\newcommand{\zeros}{\mathbf{0}}

\newcommand{\clip}[1]{\operatorname{clip}\left(#1\right)}
\newcommand{\sgn}[1]{\text{sgn}\left(#1\right)}
\newcommand{\abs}[1]{\left|#1\right|}

\newcommand{\Vector}[1]{\mathbf{#1}}
\newcommand{\Matrix}[1]{\mathbf{#1}}

\newtheorem{remark}{Remark}

\usepackage{soul}
\usepackage{tikz}
\usetikzlibrary{calc}

\makeatletter
\newif\if@anonymize

\@anonymizetrue    

\if@anonymize
  \newcommand{\highlight@DoHighlight}{
    \fill [outer sep = -15pt, inner sep = 0pt, color=black]
          ($(begin highlight)+(0,8pt)$) rectangle ($(end highlight)+(0,-3pt)$) ;
  }

  \newcommand{\highlight@BeginHighlight}{
    \coordinate (begin highlight) at (0,0) ;
  }

  \newcommand{\highlight@EndHighlight}{
    \coordinate (end highlight) at (0,0) ;
  }

  \newdimen\highlight@previous
  \newdimen\highlight@current
  \newlength{\item@width}

  \DeclareRobustCommand*\anonymize{%
    \SOUL@setup
    \def\SOUL@preamble{%
      \begin{tikzpicture}[overlay, remember picture]
        \highlight@BeginHighlight
        \highlight@EndHighlight
      \end{tikzpicture}%
    }%
    \def\SOUL@postamble{%
      \begin{tikzpicture}[overlay, remember picture]
        \highlight@EndHighlight
        \highlight@DoHighlight
      \end{tikzpicture}%
    }%
    \def\SOUL@everyhyphen{%
      \discretionary{%
        \SOUL@setkern\SOUL@hyphkern
        \SOUL@sethyphenchar
        \tikz[overlay, remember picture] \highlight@EndHighlight ;%
      }{%
      }{%
        \SOUL@setkern\SOUL@charkern
      }%
    }%
    \def\SOUL@everyexhyphen##1{%
      \SOUL@setkern\SOUL@hyphkern
      \settowidth{\item@width}{##1}%
      \makebox[\item@width]{}%
      \discretionary{%
        \tikz[overlay, remember picture] \highlight@EndHighlight ;%
      }{%
      }{%
        \SOUL@setkern\SOUL@charkern
      }%
    }%
    \def\SOUL@everysyllable{%
      \begin{tikzpicture}[overlay, remember picture]
        \path let \p0 = (begin highlight), \p1 = (0,0) in \pgfextra
          \global\highlight@previous=\y0
          \global\highlight@current =\y1
        \endpgfextra (0,0) ;
        \ifdim\highlight@current < \highlight@previous
          \highlight@DoHighlight
          \highlight@BeginHighlight
        \fi
      \end{tikzpicture}%
      \settowidth{\item@width}{\the\SOUL@syllable}%
      \makebox[\item@width]{}%
      \tikz[overlay, remember picture] \highlight@EndHighlight ;%
    }%
    \SOUL@
  }
\else
  \newcommand{\anonymize}[1]{#1}
\fi
\makeatother
\newcommand{\arms}{M}

\newcommand{\direction}{\phi}

\newcommand{\dispX}{\Delta x}
\newcommand{\dispY}{\Delta y}
\newcommand{\dispPhi}{\Delta\Phi}
\newcommand{\disps}{\Delta\mathbf{p}}
\newcommand{\dispbearing}{\vartheta}

\newcommand{\fitness}{\mathsf{J}}
\newcommand{\reward}{\mathsf{R}}
\newcommand{\Jrad}{\reward_{\mathrm{rad}}}
\newcommand{\Jang}{\fitness_{\mathrm{ang}}}
\newcommand{\Jrot}{\fitness_{\mathrm{rot}}}

\newcommand{\armctrlspace}{\mathcal{U}}
\newcommand{\ctrlspace}{\armctrlspace^{\arms}}
\newcommand{\cycperm}{P}

\newcommand{\armdim}{U}
\newcommand{\armdirection}[1][k]{\direction_{#1}}

\newcommand{\shiftedctrl}[1][\gamma\raisebox{-0.1ex}{\scalebox{0.8}{\raisedtri}}\direction]{u_{#1}}

\newcommand{\directionoffset}{\delta}

\newcommand{\blendmat}[1][\directionoffset]{A_{#1}}

\newcommand{\diffparams}{\theta}

\newcommand{\querysize}{I}

\newcommand{\diffnoise}{\epsilon}

\newcommand{\acquisition}{\alpha}

\newcommand{\scalarfitness}{f}
\newcommand{\canonfitness}{\scalarfitness_{0}}
\newcommand{\boundaryfitness}{\scalarfitness_{\mathrm{b}}}

\newcommand{\fitnessset}{\mathcal{Y}}

\newcommand{\optiter}{\ell}
\newcommand{\noptiters}{L}
\newcommand{\ensemblesize}{E}

\newcommand{\diffusionhorizon}{K}

\newcommand{\diffusionstep}{k}
\newcommand{\noisedctrl}[1][\diffusionstep]{u_{#1}}

\newcommand{\denoiser}[1]{\diffnoise_{#1}}

\newcommand{\positions}{\Vector{x}}

\newcommand{\orientation}{\Matrix{Q}}
\newcommand{\strain}{\epsilon}

\newcommand{\shears}{\bm{\nu}}

\newcommand{\angularVelocities}{\bm{\omega}}

\newcommand{\internalforce}{n}
\newcommand{\internalforces}{\Vector{n}}

\newcommand{\internalcouples}{\Vector{m}}
\newcommand{\externalforces}{\Vector{f}}
\newcommand{\externalcouples}{\Vector{c}}

\newcommand{\secondMomentOfInertia}{\Matrix{I}}

\newcommand{\muscle}{\text{m}}

\newcommand{\dataset}{\mathcal{D}}

\newcommand{\lab}[1]{\bar{#1}}

\newcommand{\Lossfunc}{\mathcal{L}}

\newcommand{\musclepositions}[1][]{\positions^{\ifthenelse{\equal{#1}{}}{\muscle}{#1}}}
\newcommand{\musclerelativepositions}[1][]{\Vector{\gamma}^{\ifthenelse{\equal{#1}{}}{\muscle}{#1}}}

\newcommand{\musclelength}[1][]{l^{\ifthenelse{\equal{#1}{}}{\muscle}{#1}}}
\newcommand{\musclestrain}[1][]{\strain^{\ifthenelse{\equal{#1}{}}{\muscle}{#1}}}
\newcommand{\muscleshears}[1][]{\shears^{\ifthenelse{\equal{#1}{}}{\muscle}{#1}}}
\newcommand{\muscletangent}[1][]{\Vector{t}^{\ifthenelse{\equal{#1}{}}{\muscle}{#1}}}
\newcommand{\muscleforce}[1][]{\internalforce^{\ifthenelse{\equal{#1}{}}{\muscle}{#1}}}
\newcommand{\maxmusclestress}[1][]{\sigma^{\ifthenelse{\equal{#1}{}}{\muscle}{#1}}}
\newcommand{\maxmuscleforce}[1][]{\internalforce^{\ifthenelse{\equal{#1}{}}{\muscle}{#1}}_\text{max}}
\newcommand{\muscleforces}[1][]{\internalforces^{\ifthenelse{\equal{#1}{}}{\muscle}{#1}}}
\newcommand{\musclecouples}[1][]{\internalcouples^{\ifthenelse{\equal{#1}{}}{\muscle}{#1}}}
\newcommand{\muscleactivation}[1][]{a^{\ifthenelse{\equal{#1}{}}{\muscle}{#1}}}
\newcommand{\staticmuscleactivation}[1][]{\alpha^{\ifthenelse{\equal{#1}{}}{\muscle}{#1}}}

\newcommand{\musclestoredenergy}[1][]{W^{\ifthenelse{\equal{#1}{}}{\muscle}{#1}}}

\newcommand{\hillsmodel}{h}

\newcommand{\TM}{\text{TM}}
\newcommand{\LM}[1][]{\text{LM}{\ifthenelse{\equal{#1}{}}{}{_{#1}}}}
\newcommand{\OM}[1][]{\text{OM}{\ifthenelse{\equal{#1}{}}{}{_{#1}}}}
\newcommand{\Sucker}[1][]{\text{S}{\ifthenelse{\equal{#1}{}}{}{_{\text{#1}}}}}

\newcommand{\identity}{I}

\newcommand{\raisedtri}{%
  \raisebox{0.1ex}{\scalebox{0.8}{$\vartriangle$}}%
}

\hypersetup{
    hidelinks=true
}

\title{\LARGE \bf Diverse and Adaptable Arm Coordination for Octopus-Crawling\\
via Diffusion-Based Uncertainty-Aware Optimization
}

\DeclareRobustCommand{\IEEEauthorrefmark}[1]{\smash{\textsuperscript{\footnotesize #1}}}

\author{
  Seung Hyun Kim\IEEEauthorrefmark{1}, 
  Heng-Sheng Chang\IEEEauthorrefmark{*,1,2}, 
  Kimia Kazemi\IEEEauthorrefmark{1}, 
  Prashant G. Mehta\IEEEauthorrefmark{1,2}, 
  Mattia Gazzola\IEEEauthorrefmark{1,3,4}
}

\begin{document}
\maketitle

\let\thefootnote\relax\footnote{%
\IEEEauthorrefmark{\scriptsize1}Mechanical Science and Engineering,
\IEEEauthorrefmark{\scriptsize2}Coordinated Science Laboratory,
\IEEEauthorrefmark{\scriptsize3}Carl R. Woese Institute for Genomic Biology,
\IEEEauthorrefmark{\scriptsize4}National Center for Supercomputing Applications,
University of Illinois Urbana-Champaign. 
\IEEEauthorrefmark{\scriptsize*}CA email: hschang@illinois.edu.
Code repository will be available soon.

This work is supported by the Office of Naval Research (N00014–22–1–2569), and NSF Expedition ``Mind in Vitro'' award (IIS-2123781), and used TACC Frontera HPC allocation at the Texas Advanced Computing Center through Allocation (IBN22011).
}

\thispagestyle{empty}
\pagestyle{empty}

\vspace*{-2\baselineskip}

\begin{abstract}

  Octopus crawling motivates soft robots that exploit redundancy, yet discovering and organizing diverse coordination modes for adaptation remains challenging. To address this, we introduce a Diffusion-based Uncertainty-aware Optimization (DUO) algorithm that learns demonstration-free crawling controllers for a simulated, muscle-actuated \emph{CyberOctopus}. This work represents the first application of diffusion-based control to soft multi-arm robots in contact-rich simulations. By embedding a variety of locomotion behaviors within a shared control distribution, this approach enables the simulated octopus to navigate dynamic physical constraints, demonstrating that learned coordination diversity inherently facilitates robust adaptation. The main contributions include: (i) a symmetry-structured policy representation that folds radially equivalent controllers into a canonical directional sector, (ii) an online black-box optimization strategy, the DUO algorithm, that discovers and retains diverse coordination modes, and (iii) a control editing technique that adapts existing controllers to novel actuator constraints without retraining. These results show how learned coordination diversity makes motor abundance a practical resource for adaptation in soft multi-arm robots.

\end{abstract}





\section{INTRODUCTION}

Octopus crawling offers both inspiration and a challenge for soft-arm robotics.
Through body compliance, distributed muscular actuation, and coordinated recruitment of redundant arms~\cite{Grasso2008suckerarm,Kennedy2020arm}, octopuses crawl without a fixed gait or rhythmic motor pattern~\cite{Levy2015armcoordination,Levy2017crawl}.
They appear to exploit their excess physical degrees of freedom as a resource for diverse behaviors and survival in natural environments~\cite{bennice2025octopus,Margheri2012histology}.
This interplay between motor abundance and compliant body mechanics~\cite{Hauser2011morphological} motivates supporting a range of coordination strategies for adaptation under unexpected actuation and environmental constraints.

Realizing this requires exploring a high-dimensional control space, an already challenging endeavor, which is further complicated by the nonlinear mechanics and contact dynamics of soft arms.
Existing approaches pursue behavioral diversity through explicit archives~\cite{Pugh2016QD,nilsson2021Elites}, novelty-driven exploration~\cite{lehman2011novelty}, and hierarchical structures~\cite{shih2023hierarchical,allard2023hierarchical}.
As system dimensionality scales, engineering these behavioral descriptors becomes a major bottleneck~\cite{hedayatian2026autoqd}. 
The central challenge, therefore, is to learn an implicit representation that natively encapsulates behavioral diversity without requiring explicit parameterization.

To learn such a representation directly from data~\cite{bengio2013RepLearn}, we use a conditional diffusion model that captures a multimodal distribution within a shared latent space~\cite{chi2024diffusionpolicy}. 
While such models usually demand extensive prior datasets, this limitation is bypassed by constructing a dataset dynamically via online black-box (simulator-based) optimization~\cite{krishnamoorthy2023DDOM}. 
However, evaluating candidate controls in simulation is costly; therefore, we guide our exploration of the control space by integrating predictive uncertainty~\cite{wu2024diffusionbbo,yun2025dibo}, ensuring our limited budget is spent testing only the informative candidates. 
This acquisition strategy underpins our Diffusion-based Uncertainty-aware Optimization (DUO) algorithm.

\begin{figure}[t!]
    \vspace{5pt}
    \vspace*{-\baselineskip}
    \centering
    \includegraphics[width=0.9\linewidth]{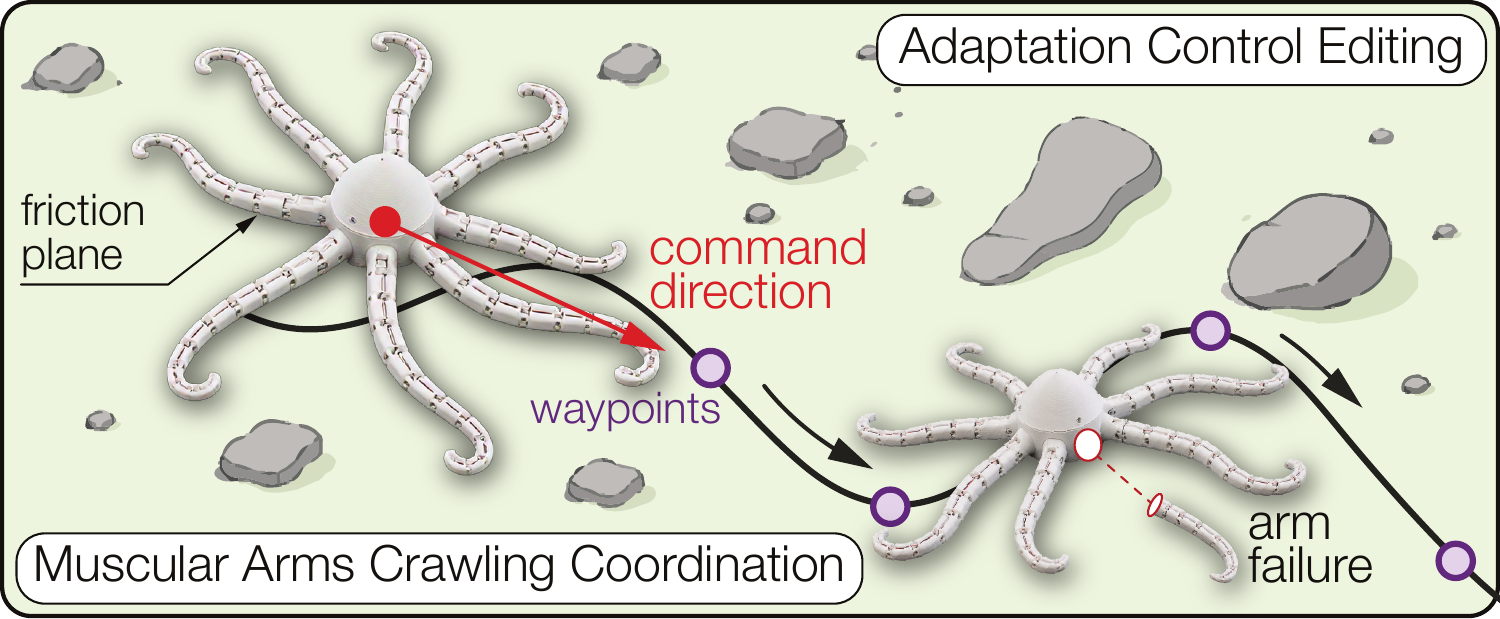}
    \vspace*{-5pt}
    \caption{\textbf{CyberOctopus.}
    The eight-arm soft-body model crawls on a frictional plane following waypoint sequences, requiring coordinated distributed actuation in a hyper-redundant control space.
    A diffusion model is chosen to handle a variety of feasible coordination modes, and the trained model can then be leveraged for adaptive control editing when an arm fails or surface friction changes, without requiring retraining.
    }
    \label{fig1:concept}
    \vspace*{-15pt}
\end{figure}

We evaluate the proposed algorithm using a simulated muscular \emph{CyberOctopus} ~(\Cref{fig1:concept}) with eight radially arranged continuum arms~\cite{shih2023hierarchical} and distributed muscle actuation~\cite{chang2023energyshaping,tekinalp2024topology}.
Because no prior demonstration data exists for this complex morphology, we rigorously test the algorithm's ability to learn full-direction crawling entirely from dynamic simulation. 
We then assess the robustness of the resulting controller by subjecting the system to previously unseen physical constraints, such as actuator failures or sudden changes in surface friction, to evaluate its capacity for adaptive locomotion.
Our contributions are threefold:
\begin{itemize}
    \item \textbf{Symmetry folding.} 
    We reduce the search direction to a single forward canonical sector by exploiting the system's rotational symmetry, utilizing interpolation and boundary evaluation to enable full-direction crawling.

    \item \textbf{Diversity-oriented search.}
    We introduce the DUO algorithm, which actively guides exploration to discover and retain a diverse family of effective coordination modes within a limited simulation budget.

    \item \textbf{Constrained reuse.}
    We demonstrate zero-shot adaptation through control editing. 
    By leveraging the retained coordination modes, the system successfully recovers locomotion under novel physical limitations without requiring any retraining of the diffusion model.
\end{itemize}


\section{RELATED WORK}
\label{sec:related-work}

Controller discovery for soft multi-arm crawling can be formulated as black-box optimization, with the locomotion performance objective obtained through costly simulation evaluations. 
The redundancy of multi-arm systems further motivates discovering distinct coordination patterns that achieve comparable performance. 
Accordingly, we review quality--diversity methods for realizing diverse, high-performing solutions and diffusion-based optimization methods for representing multimodal solution distributions.

\subsubsection{Quality--diversity (QD)}
These methods seek collections of high-performing solutions across a \emph{behavior space} defined by descriptors~\cite{Pugh2016QD,nilsson2021Elites}.
Classical approaches use handcrafted descriptors to organize an archive;
grid-based archives retain the highest-performing solution in each occupied cell. 
This organization supports the discovery and reuse of behavioral repertoires in robot learning and adaptation~\cite{allard2023hierarchical}.

The choice of descriptors dictates which behavioral variations are preserved within the archive.
For multi-arm crawling, final-displacement descriptors can group distinct coordination patterns into the same cell, while detailed arm-deformation descriptors increase representation and archive--design complexity. 
AutoQD~\cite{hedayatian2026autoqd} addresses the need for manual descriptor design by embedding policy occupancy measures and projecting them into a low-dimensional behavior space for QD optimization. 
These learned descriptors guide archive-based search toward behaviorally distinct solutions without predefined behavioral descriptors.

\subsubsection{Diffusion-based optimization}
Diffusion models learn complex, multimodal distributions through iterative denoising~\cite{ho2020denoising}, enabling expressive action distributions in robotic control~\cite{chi2024diffusionpolicy}. 
In optimization, Diffusion Evolution~\cite{zhang2025diffusionevolution} connects denoising with evolutionary optimization, while DDOM~\cite{krishnamoorthy2023DDOM} learns objective-conditioned distributions from fixed datasets. 
Building on these offline methods, online extensions can actively evaluate candidates during search.

DiffBBO~\cite{wu2024diffusionbbo} and DiBO~\cite{yun2025dibo} alternate candidate generation, evaluation, and model updates for online black-box optimization.
DiffBBO uses ensemble uncertainty to select target objective values for conditional generation, whereas DiBO combines surrogate-guided posterior inference with local refinement. 
These approaches highlight the importance of acquisition design under limited evaluation budgets: emphasizing predicted performance can concentrate search, while exploration is needed to discover alternative solutions.

\begin{figure*}[t!]
    \vspace{5pt}
    \centering
    \includegraphics[width=\linewidth]{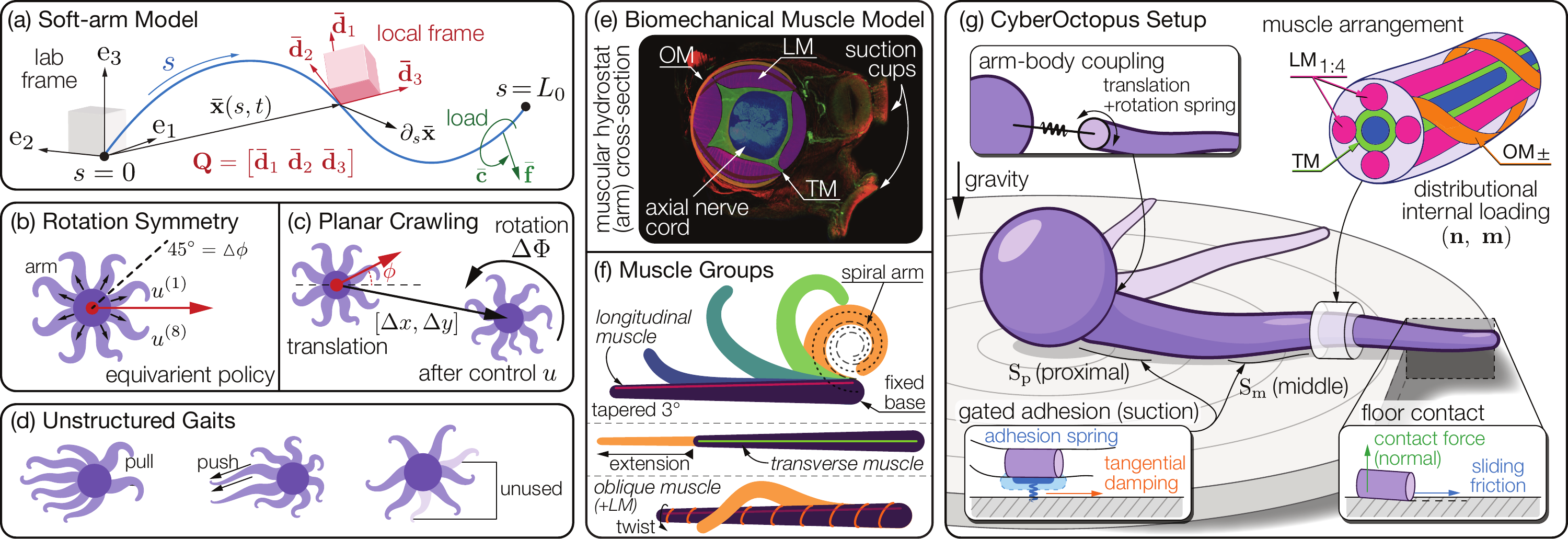}
    \caption{\textbf{Cyber-octopus model and actuation.}
    (a) Cosserat-rod soft-arm model. 
    (b) Eightfold radial symmetry.
    (c) Control-induced translation and body rotation.
    (d) Arm coordination through pulling, pushing, and alternating recruitment.
    (e) Muscular-hydrostat cross-section: longitudinal (LM), transverse (TM), and oblique (OM) muscles, axial nerve cord, and suction cups.
    (f) Muscle-driven bending (LM), extension (TM), and twisting (OM and LM);
    tapered arms and muscle paths follow a spiral construction~\cite{Wang2023SpiRobs}.
    (g) Muscle arrangement, arm-body coupling, internal loading, and floor contact.
    Within the capture distance, sucker adhesion attracts the arm to the floor while tangential damping suppresses slip.}
    \label{fig2:setup}
    \vspace{-15pt}
\end{figure*}

\section{PROBLEM FORMULATION}
\label{sec:problem}

We formulate planar crawling control as an optimization problem: discover a family of controls that achieves stable crawling across the full range of directions, while penalizing net body rotation to preserve heading alignment.

\subsubsection{Control evaluation}
\label{sec:controller_evaluation}
Let $u\in\ctrlspace$ denote a control for a robot with $\arms$ arms (\Cref{fig2:setup}b). 
The control to the $m$-th arm is $u^{(m)}\in\armctrlspace=[-1,1]^{\armdim}$, and
\begin{equation*}
    u\defined\bigl[u^{(1)},\ldots,u^{(\arms)}\bigr]
    \in\ctrlspace=[-1,1]^{\arms\times\armdim}.
\end{equation*}
Applying a control $u$ yields a body rotation $\dispPhi(u)$ and a planar displacement $\disps(u)=[\dispX(u),\dispY(u)]$ with $\dispbearing(u)=\arctan2(\dispY(u),\dispX(u))$.
For a commanded direction $\direction\in[0,2\pi]$,  the crawling performance is quantified by
\begin{equation*}
    \begin{aligned}
        \Jrad(u;\direction)
        &\defined \cos\direction\,\dispX(u)+\sin\direction\,\dispY(u),\\
        \Jang(u;\direction)
        &\defined \abs{\arctan2\!\left(
            \sin(\dispbearing(u)-\direction),
            \cos(\dispbearing(u)-\direction)\right)},\\
        \Jrot(u)&\defined\abs{\dispPhi(u)}.
    \end{aligned}
\end{equation*}
$\Jrad$ is progress along the direction $\direction$, $\Jang$ is the angular deviation, and $\Jrot$ is the magnitude of the body rotation. 

\subsubsection{Control discovery as an optimization}
For a fixed commanded direction $\direction$, control discovery is formulated as optimization of the following objective:
\begin{equation}
    \max_{u\in\ctrlspace}\Jrad(u;\direction)
    -\frac{\lambda}{2}\left[\Jang(u;\direction)+\Jrot(u)\right],
    \label{eq:crawling_objective}
\end{equation}
where $\lambda>0$ scales the angular penalties into the displacement units of
$\Jrad$. 
The objective rewards translation along the commanded direction $\direction$ while discouraging directional error and net body rotation. 
While incorporating commanded body rotation is a natural extension, limiting the formulation to pure translation provides a clearer baseline for evaluating crawling performance. 
Thus, we scope our study to translational motion and defer rotational control to future work.

\section{METHOD}
\label{sec:method}


\subsection{Radial Symmetry Reduction}
\label{sec:radial-symmetry}

\subsubsection{Rotational equivariance}
Solving \eqref{eq:crawling_objective} independently for every $\direction$ would require a continuum of direction-specific searches. 
We instead use the rotational symmetry of the robot to reduce the search space.
Let $\raisedtri\direction\defined\nicefrac{2\pi}{\arms}$ be the angle between adjacent arms, $\cycperm$ the cyclic permutation:
\begin{equation*}
    \begin{aligned}
        \cycperm \defined \begin{bmatrix}\zeros&1 \\ \identity_{\arms-1}&\zeros\end{bmatrix},~ \shiftedctrl \defined \cycperm^{\gamma}u, ~\forall \gamma\in\{0,1,\dots,\arms-1\}.
    \end{aligned}
\end{equation*}
Thus, $\cycperm$ maps $[u^{(1)},\ldots,u^{(\arms)}]$ to $[u^{(\arms)},u^{(1)},\ldots,u^{(\arms-1)}]$ and satisfies $\cycperm^{\arms}=\identity_\arms$ and $\cycperm^{-1}=\cycperm^{\arms-1}$.
The shifted control $\shiftedctrl$ rotates a control assignment $u$ by $\gamma$ arms.

Exploiting this symmetry, the permutation rotates crawling direction while preserving direction-relative performance:
\begin{equation*}
    \begin{aligned}
        \Jrad(\shiftedctrl;\direction-\gamma\raisedtri\direction)&=\Jrad(u;\direction),\\
        \Jang(\shiftedctrl;\direction-\gamma\raisedtri\direction)&=\Jang(u;\direction),\\
        \Jrot(\shiftedctrl)&=\Jrot(u),
    \end{aligned}
    \quad \forall \gamma\in\{0,\ldots,\arms-1\}.
\end{equation*}
It is therefore sufficient to search the canonical sector $[\nicefrac{-\raisedtri\direction}{2},\nicefrac{\raisedtri\direction}{2}]$; 
the other sectors follow by cyclic permutation. 
This gives exact controls at the discrete symmetry directions, and intermediate directions require interpolation.

\subsubsection{Full-direction control}
We write the commanded direction as $\direction= \gamma\raisedtri\direction+\directionoffset$ with $\gamma\in\{0,\ldots,\arms-1\}$ and $\directionoffset\in[\nicefrac{-\raisedtri\direction}{2},\nicefrac{\raisedtri\direction}{2}]$, blending the adjacent arm assignments using
\begin{equation*}
    u_{\direction}\defined\blendmat\cycperm^\gamma u,\quad \blendmat\defined\left(1-\tfrac{\abs{\directionoffset}}{\raisedtri\direction}\right)\identity_\arms + \tfrac{\abs{\directionoffset}}{\raisedtri\direction}\cycperm^{\hspace{0.1em}\sgn{\directionoffset}}.
\end{equation*}
Because both $\cycperm$ and $\blendmat$ map $\ctrlspace$ to itself,
$u_{\direction}$ remains a valid control.
This construction produces a directional control based on one canonical control for all $\direction\in[0,2\pi]$.

\subsubsection{Boundary-robust objective}
Rotational equivariance does not determine performance between symmetry-related directions. 
We therefore evaluate each control both at the canonical direction $\directionoffset=0$ and at $\directionoffset=\nicefrac{\raisedtri\direction}{2}$, the boundary of the canonical sector. 
The boundary at the other side $\nicefrac{-\raisedtri\direction}{2}$ is equivalent by the cyclic symmetry, so one additional evaluation suffices.

Define the canonical and boundary radial objectives as
\begin{equation*}
    \canonfitness(u)\defined\Jrad(u;0),
    \quad
    \boundaryfitness(u)\defined
    \Jrad\!\left(\blendmat[\nicefrac{-\raisedtri\direction}{2}]u;
    \tfrac{\raisedtri\direction}{2}\right).
\end{equation*}
The objective function combines these terms and applies the directional and rotational penalties at both the canonical and boundary directions:
\begin{equation}
    \begin{aligned}
        \scalarfitness(u)&\defined\left[\canonfitness(u)+\boundaryfitness(u)\right]-\frac{\lambda}{2}\left[\Jang(u;0)+\Jrot(u)\right]\\
        &\quad -\frac{\lambda}{2}\left[\Jang(\blendmat[\nicefrac{-\raisedtri\direction}{2}]u;\tfrac{\raisedtri\direction}{2})+\Jrot(\blendmat[\nicefrac{-\raisedtri\direction}{2}]u)\right]
    \end{aligned}
    \label{eq:objective_function}
\end{equation}
with $\lambda>0$ as the angular penalty weight.
Weighting the canonical and boundary terms equally prevents an optimized control from trading boundary performance for center canonical direction performance.
The reformulated optimization problem \eqref{eq:crawling_objective} with the radial-symmetry reduction becomes
\begin{equation}
    \max_{u\in\ctrlspace}~\scalarfitness(u),
    \label{eq:scalar_fitness}
\end{equation}
where $y=\scalarfitness(u)$ is the objective value for control $u\in\ctrlspace$.

\begin{remark}
   Radial symmetry reduces the search dimensionality to a single canonical sector. 
   However, solving \eqref{eq:scalar_fitness} still presents two challenges:
    \begin{enumerate}
        \item The remaining high-dimensional actuation space $\ctrlspace$.
        \item The high computational cost of objective $\scalarfitness$ evaluation.
    \end{enumerate}
    We address these challenges in the following subsection.
    \label{remark:challenges}
\end{remark}

\begin{figure*}[t!]
    \vspace{5pt}
    \centering
    \includegraphics[width=\linewidth]{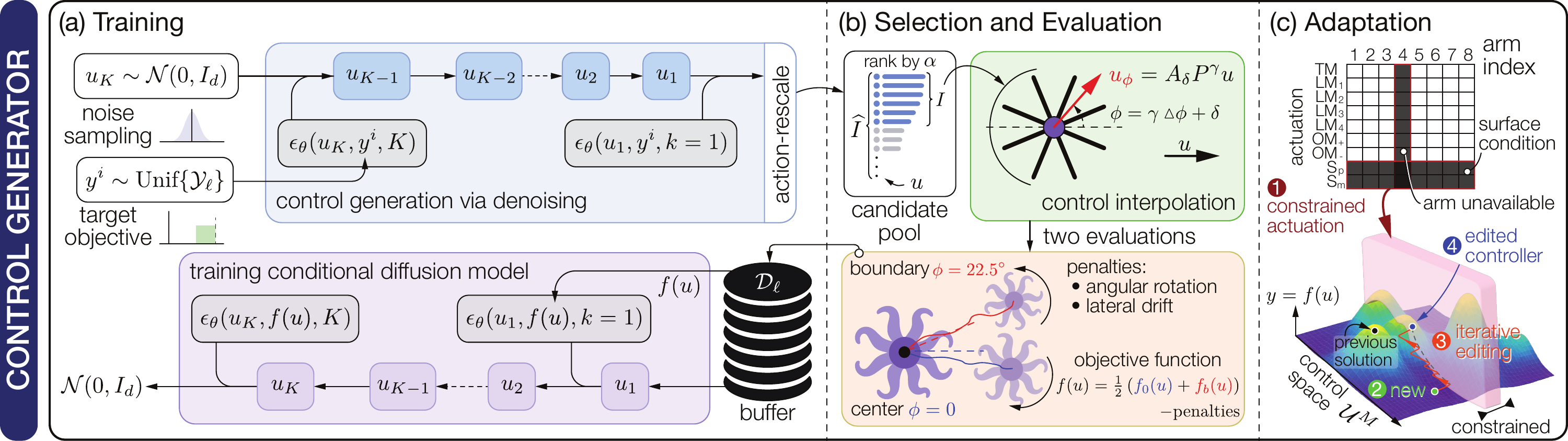}
    \caption{ \textbf{Control discovery and constrained reuse.}
        (a) Control generation and denoiser training.
        (b) Candidate selection and evaluation. 
        Candidates are ranked by acquisition score $\acquisition$, and the top $\querysize$ candidates are evaluated.
        Simulating the crawling motion in both the center and boundary directions yields the objective values for the next iteration.
        (c) Control reuse by editing.
        Shaded grid entries illustrate constrained actuation coordinates.
        The landscape depicts a schematic objective, and the pink hyperplane denotes the constrained control subspace.
        Through repeated noising and denoising, a previous control can be adjusted to satisfy the constraints while improving performance, without retraining.
    }
    \label{fig3:method}
    \vspace{-15pt}
\end{figure*}

\subsection{Diffusion-based Uncertainty-aware Optimization (DUO)}
\label{sec:diffusion-formulation}

\subsubsection{Diffusion-based inverse model}
To address the challenge of searching in a high-dimensional control space $\ctrlspace\cong[-1,1]^d$ with $d=\arms\armdim$, 
we propose candidate controls by sampling from a trained inverse model $p_{\diffparams}:\reals\to\mathcal{P}(\ctrlspace)$, a conditional distribution that predicts control distributions given an objective value from $\scalarfitness:\ctrlspace\to\reals$.
For diffusion modeling, we identify each control $u$ with its vectorization \(\operatorname{vec}(u)\), and reshape generated samples before simulation.

A denoising diffusion probabilistic model (DDPM)~\cite{ho2020denoising} learns the conditional distribution $p_{\diffparams}(\cdot\mid y)$, where $\diffparams$ denotes its trainable parameters and $y$ the requested objective value of $\scalarfitness$.
For a clean control $\noisedctrl[0]\in[-1,1]^d$, the diffusion process produces a sequence of noisy controls $\noisedctrl[1],\ldots,\noisedctrl[\diffusionhorizon]$ through
\begin{equation*}
    \noisedctrl = \clip{\sqrt{\bar a_{\diffusionstep}}\,\noisedctrl[0] + \sqrt{1-\bar a_{\diffusionstep}}\,\diffnoise_{\diffusionstep}},
    \quad \diffnoise_{\diffusionstep}\sim\mathcal{N}(0,I_d),
\end{equation*}
where $\diffusionstep\in\{1,\ldots,\diffusionhorizon\}$ indexes the $\diffusionhorizon$ diffusion steps, $\bar a_{\diffusionstep}\in(0,1)$ is the monotonically decreasing cumulative signal-retention factor prescribed by the noise variance schedule.
The clipping operator restricts the noisy control to $[-1,1]^d$.
The DDPM trains a noise predictor $\denoiser{\diffparams}$ minimizing
\begin{equation}
    \Lossfunc(\diffparams;\dataset)
    \defined\mathbb{E}\!\left[
    \left\|\diffnoise-\denoiser{\diffparams}
    (\noisedctrl,y,\diffusionstep)\right\|_2^2\right],
    \label{eq:ddom_loss}
\end{equation}
where the expectation is over $(\noisedctrl[0],y)\sim \operatorname{Unif}(\dataset)$, a uniform step $\diffusionstep$, and Gaussian noise $\diffnoise$. Here, $\dataset$ contains the evaluated control-objective pairs $(u,\scalarfitness(u))$.

For a fixed target objective $y$, sampling begins with $\noisedctrl[\diffusionhorizon]\sim\mathcal{N}(0,I_d)$ and proceeds through reverse steps $\diffusionstep=\diffusionhorizon,\ldots,1$.
At each step $\diffusionstep$, the denoiser computes the conditional mean, and the next control is sampled as~\cite{ho2020denoising}
\begin{equation*}
    \begin{aligned}
        \bar{u}_{\diffparams,\diffusionstep}
        &=\sqrt{\tfrac{\bar{a}_{\diffusionstep-1}}{\bar{a}_{\diffusionstep}}}\left[
        \noisedctrl-\tfrac{\tilde{u}_{\diffusionstep}^2}{\sqrt{1-\bar a_{\diffusionstep}}}
        \denoiser{\diffparams}(\noisedctrl,y,\diffusionstep)\right],\\
        \noisedctrl[\diffusionstep-1]
        &=\clip{\bar{u}_{\diffparams,\diffusionstep}+\tilde{u}_{\diffusionstep}\diffnoise_{\diffusionstep}},\quad\tilde{u}_{\diffusionstep}=\sqrt{1-\tfrac{\bar{a}_{\diffusionstep}}{\bar{a}_{\diffusionstep-1}}},
    \end{aligned}
\end{equation*}
where fresh noise vectors $\diffnoise_{\diffusionstep}\sim\mathcal{N}(0,I_d)$ are drawn independently of one another for $\diffusionstep>1$, and $\diffnoise_1=0$ makes the final step return the denoised mean, following Algorithm~2 of~\cite{ho2020denoising}.
After $\diffusionhorizon$ reverse steps, the final control $u=\noisedctrl[0]$ is a sample from $p_{\diffparams}(\cdot\mid y)$.
Resampling with fresh noise yields alternative controls for the same target objective $y$.

This inverse model replaces direct search in~\eqref{eq:scalar_fitness} with target-selection to maximize expected control performance:
\begin{equation}
    \max_{y\in\fitnessset}~\mathbb{E}_{u\sim p_{\diffparams}(\cdot\mid y)}[\scalarfitness(u)],
    \label{eq:ddom_acquisition}
\end{equation}
where $\fitnessset\subseteq\reals$ is the candidate target objective set.
The design of $\fitnessset$ and the remaining hyperparameters is specified in \Cref{sec:experiments}.

Solving \eqref{eq:ddom_acquisition} requires a representative dataset $\dataset$. 
When no pre-existing demonstration data is available, the diffusion model must be trained by iteratively evaluating proposed controls in simulation and updating the dataset with the resulting objective values as $\dataset_{\optiter+1}=\dataset_{\optiter}\cup\{(u^i,\scalarfitness(u^i))\}_{i=1}^{\querysize}$ with $\querysize$ new evaluations at iteration $\optiter$.
However, evaluating the objective function $\scalarfitness$ is computationally expensive, and therefore simulating every control candidate generated by the inverse model is intractable.
To efficiently allocate the evaluation budget, we employ an acquisition strategy that uses predicted performance to select a limited, informative subset of candidates $\fitnessset_{\optiter}$ for simulation.

\begin{figure}[t] 
    \vspace*{5pt}  
    \hrule        
    \vspace{5.25pt}  
    \captionof{algorithm}{\textbf{Diffusion-based Uncertainty-aware Optimization (DUO)}}
    \label{alg:diffusion-training}
    \vspace{-4.25pt}
    \hrule        
    \vspace{2pt}
    \begin{algorithmic}[1]
        \Require objective $\scalarfitness$; initial dataset $\dataset_0$; iteration count $\noptiters$; model number $E,J$; batch size $\querysize$; proposal count $\widehat{\querysize}>\querysize$
        \Ensure trained conditional generator $p_{\diffparams}(u\mid y)$
        \For{$\optiter=0,\ldots,\noptiters-1$}
            \State Update inverse function generators with $\dataset_{\optiter}$:
            \Statex \hspace{\algorithmicindent} train diffusion models $\{p_{\diffparams_e}(u\mid y)\}_{e=1}^{\ensemblesize}$ with~\eqref{eq:ddom_loss}
            \State Update forward objective estimators with $\dataset_{\optiter}$:
            \Statex \hspace{\algorithmicindent} train surrogates $\{\hat{\scalarfitness}_{\psi_j}(u)\}_{j=1}^{J}$ with MSE regression
            \State Sample candidate targets $\{y^i\sim\operatorname{Unif}\{\fitnessset_{\optiter}\}\}_{i=1}^{\widehat{\querysize}}$
            \State Sample $\{u^i\sim p_{\diffparams_e}(u\mid y^i):e\sim\operatorname{Unif}\{1,\ldots,\ensemblesize\}\}_{i=1}^{\widehat{\querysize}}$
            \State Rank $\{(u^i, y^i)\}_{i=1}^{\widehat{\querysize}}$ using acquisition score \eqref{eq:acquisition_final}
            \State Retain and relabel the top candidates as $\{u^i\}_{i=1}^{\querysize}$
            \State Evaluate $\{\scalarfitness(u^i)\}_{i=1}^{\querysize}$ with the dynamic simulator
            \State $\dataset_{\optiter+1}\leftarrow\dataset_{\optiter}\cup\{(u^i,\scalarfitness(u^i))\}_{i=1}^{\querysize}$
        \NoNumberEndFor
        \State $p_{\diffparams}(u\mid y) \leftarrow p_{\diffparams_e}(u\mid y)$, $e\sim\operatorname{Unif}\{1,\ldots,\ensemblesize\}$
    \end{algorithmic}
    \vspace{2pt}
    \hrule        
    \vspace*{-15pt}
\end{figure}

\subsubsection{Uncertainty-aware batch acquisition}
To address the challenge of the costly dynamic simulation, we extract a highly informative evaluation batch from a large pool of generated proposals. 
Candidates are scored by combining their target reliability, estimated via the epistemic uncertainty of a diffusion model ensemble, with their predicted control performance, evaluated through an ensemble of surrogate models of objective function $f$. 
We then simulate only the top-ranked batch and append the resulting ground-truth evaluations to the dataset for the next optimization iteration.

Following DiffBBO~\cite{wu2024diffusionbbo}, we train $\ensemblesize$ independently initialized diffusion models with parameters $\{\diffparams_e\}_{e=1}^{\ensemblesize}$.
For $e\sim\operatorname{Unif}\{1,\ldots,\ensemblesize\}$, define the target epistemic uncertainty:
\begin{equation*}
    \Delta_{\mathrm{epi}}(y;\dataset_{\optiter})\defined\operatorname{Var}_{e}\!\left[
    \mathbb{E}_{u\sim p_{\diffparams_e}(\cdot\mid y)}\left[\tfrac{\|u\|_2}{\sqrt{d}}\right]
    \right].
\end{equation*}
The inner expectation estimates one model's candidate norm, and the outer variance measures disagreement among models.
The logarithmic implementation of uncertainty-aware exploration (UaE)~\cite{wu2024diffusionbbo} assigns the target score
\begin{equation*}
    \acquisition^{\mathrm{target}}(y;\dataset_{\optiter})
    \defined\log\frac{\bar{y}}{\Delta_{\mathrm{epi}}(y;\dataset_{\optiter})},\quad \bar{y}\defined\frac{y-y_{\min,\optiter}}{y_{\max,\optiter}-y_{\min,\optiter}},
\end{equation*}
where $y_{\min,\optiter}$ and $y_{\max,\optiter}$ are the minimum and maximum objective values in $\dataset_{\optiter}$.
This score $\acquisition^{\mathrm{target}}$ favors high target objectives with low ensemble model disagreement.

To distinguish control candidates $u$ generated at the same target objective value $y$, we complement target reliability $\acquisition^{\mathrm{target}}$ with predictions of individual control performance $\acquisition^{\mathrm{control}}$.
Following DiBO~\cite{yun2025dibo}, we train $J$ independently initialized neural-network surrogates $\{\hat{\scalarfitness}_{\psi_j}\}_{j=1}^{J}$ by mean-squared-error regression on $\dataset_{\optiter}$.
Here, $\psi_j$ denotes the trainable parameters of surrogate $\hat{\scalarfitness}_{\psi_j}$, whose output estimates $\scalarfitness$.
For $j\sim\operatorname{Unif}\{1,\ldots,J\}$, define the predictive moments
\begin{equation*}
        \mu(u;\dataset_{\optiter})\defined\mathbb{E}_{j}[\hat{\scalarfitness}_{\psi_j}(u)], \quad \sigma^2(u;\dataset_{\optiter})\defined\operatorname{Var}_{j}[\hat{\scalarfitness}_{\psi_j}(u)].
\end{equation*} 
The upper-confidence-bound (UCB) score of each candidate control $u$ is then formulated as
\begin{equation*}
    \acquisition^{\mathrm{control}}(u;\dataset_{\optiter})\defined\mu(u;\dataset_{\optiter})+\kappa\sigma(u;\dataset_{\optiter}),
\end{equation*}
where $\kappa\geq0$ balances the preference for high predicted performance (exploitation) against the preference for uncertain predictions (exploration).
Both statistics depend on the current dataset $\dataset_{\optiter}$, and the UCB score is computed for each candidate control $u$ generated at its assigned target $y$.

For a candidate pair $(u,y)$, the combined score is
\begin{equation}
    \acquisition(u,y;\dataset_{\optiter})\defined\acquisition^{\mathrm{control}}(u;\dataset_{\optiter})+\eta\acquisition^{\mathrm{target}}(y;\dataset_{\optiter}),
    \label{eq:acquisition_final}
\end{equation}
where $\eta\geq0$ weights target reliability relative to control performance.
This score is used to rank candidates for selecting the most promising ones to evaluate in simulation, balancing the trade-off between high predicted performance and low uncertainty.

At each iteration, we sample $\widehat{\querysize}>\querysize$ candidate targets in $\fitnessset_{\optiter}$, where $\widehat{\querysize}$ is the proposal count and $\querysize$ the new simulation batch size.
For each candidate index $i\in\{1,\ldots,\widehat{\querysize}\}$, we draw $u^i\sim p_{\diffparams_e}(\cdot\mid y^i)$ at its assigned target $y^i\sim\operatorname{Unif}\{\fitnessset_{\optiter}\}$, with $e$ sampled uniformly from the diffusion ensemble.
Candidates receive their combined score $\acquisition(u^i,y^i;\dataset_{\optiter})$ and are ranked accordingly.
The top $\querysize$ candidates are selected for simulation, and their resulting objective values supply new pairs $\{(u^i,\scalarfitness(u^i))\}_{i=1}^{\querysize}$ to the dataset for the next iteration.
\Cref{alg:diffusion-training} summarizes the full iterative procedure, illustrated in \Cref{fig3:method}a--b.

\section{EXPERIMENT SETUP}
\label{sec:experiments}

\begin{figure*}[t!]
    \vspace{5pt}
    \centering
    \includegraphics[width=\linewidth]{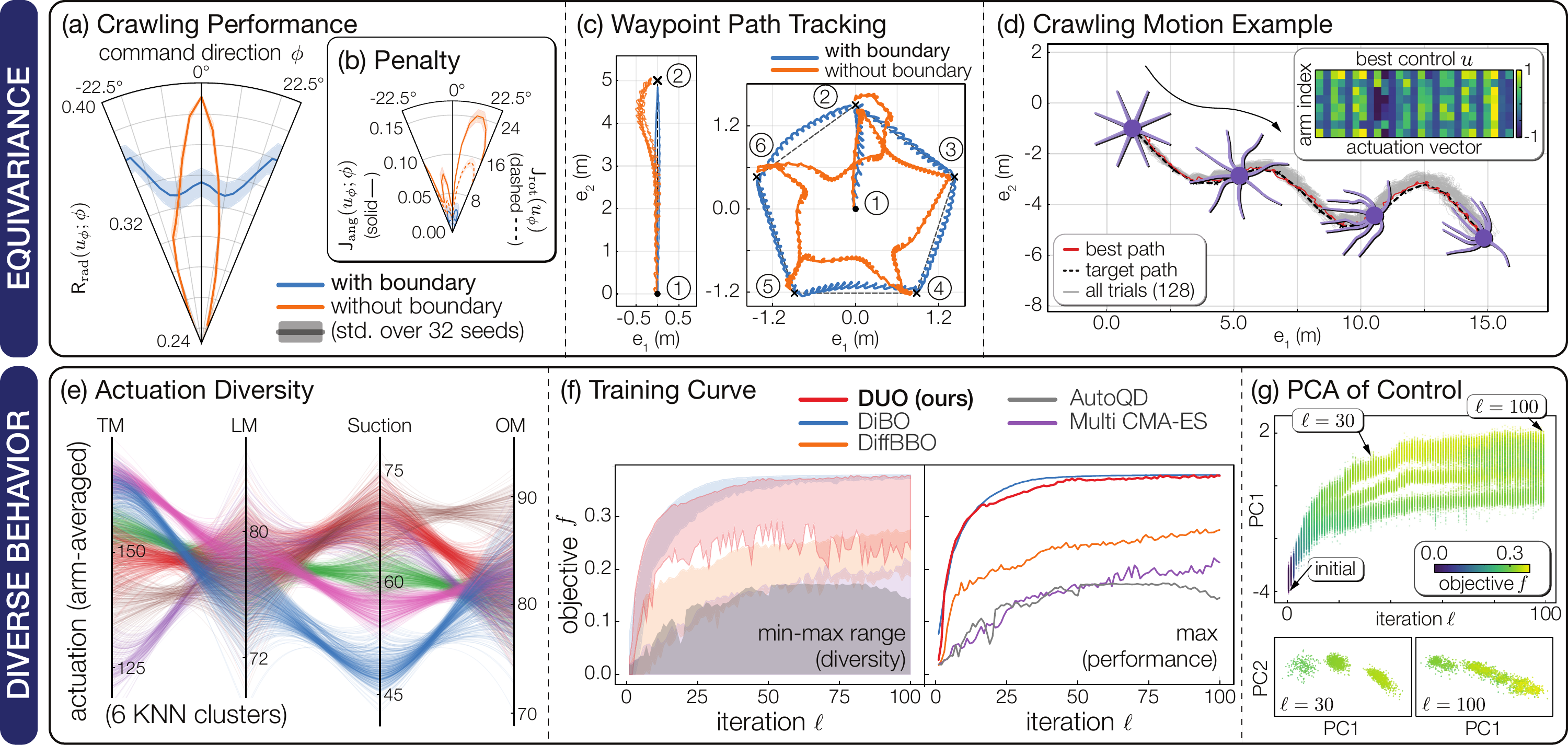}
    \caption{\textbf{Directional crawling and control diversity.}
    (a) Crawling objective over $\direction\in[-22.5^{\circ},22.5^{\circ}]$ with (blue) and without (orange) the boundary objective; shading indicates standard deviation over $32$ seeds.
    (b) Angular deviation (solid, left scale) and body rotation (dashed, right scale).
    (c) Waypoint tracking with the same color convention as (a); black dashed lines mark the commanded paths.
    (d) Crawling along a curved target path (black dashed), showing successive body configurations, trajectories from $128$ controls (gray), and best control (red); inset: best control values.
    (e) Arm-averaged time-integrated activation of TM, LM, suction, and OM recruitment across the control family. The family is clustered using the k-nearest neighbors (KNN) algorithm ($\text{k}=6$).
    (f) Per-iteration objective ranges (left) and maxima (right) for the compared methods.
    (g) PCA of the generated control distribution across training iterations, $1,800$ samples per iteration. PC1 and PC2 indicate the first and second principal components.
    }
    \label{fig4:result1}
    \vspace{-15pt}
\end{figure*}

\subsubsection{CyberOctopus model}
\label{sec:cyber-octopus}

The CyberOctopus models a soft-arm robot driven by distributed internal actuation ~(\Cref{fig2:setup}).
Its architecture is inspired by the octopus arm's muscular hydrostat~\cite{Margheri2012histology}, with deformation capabilities relevant to pneumatic, tendon-driven, and embedded soft-actuation systems~\cite{rus2015design,cianchetti2018biomedical}.
Each tapered arm $m$ is represented as a geometrically exact Cosserat rod~\cite{Antman1995elasticity,Gazzola2018Cosserat} that captures large bending, twisting, shear, and extension ~(\Cref{fig2:setup}a).
The rod is parameterized by arc length $s\in[0,L_0]$ and time $t\in[0,T]$, where $L_0$ is the arm length at rest and $T$ is the locomotion cycle duration.
Each of the $\arms$ arms connects to a central spherical mantle via translational and rotational restoring springs. 
In their resting state, the arms are spaced at equal intervals and align with the directional vectors 
\begin{equation*}
    (\cos\armdirection[m],~\sin\armdirection[m]),\quad \armdirection[m]=(m-\tfrac{1}{2})\raisedtri\direction
\end{equation*}
for all $m\in\{1,2,\hdots,\arms\}$.

Let $\lab\positions(s,t)$ and $\orientation(s,t)$ denote an arm's centerline position and material frame. Its linear and angular momentum balances are
\begin{equation}
    \begin{aligned}
        \partial_{t}(\rho\partial_{t}\lab\positions) & =\partial_{s}(\orientation\internalforces) +\lab\externalforces,\\
        \partial_{t}(\secondMomentOfInertia\angularVelocities) & =\partial_{s}(\orientation\internalcouples) +(\partial_{s}\lab\positions)\times(\orientation\internalforces) + \lab\externalcouples,
    \end{aligned}
    \label{eq:main_cosserat_dynamics}
\end{equation}
where $\rho$ and $\secondMomentOfInertia$ are the mass and second moment of inertia per unit length, respectively, and $\angularVelocities$ is the angular velocity.
The pairs $(\internalforces,\internalcouples)$ and $(\lab\externalforces,\lab\externalcouples)$ denote the internal force and couple resultants and the external force and couple densities, respectively.
External loads comprise gravity, compliant normal contact, anisotropic kinetic Coulomb friction, arm-to-arm contact, and sucker adhesion~\cite{tekinalp2024topology,kim2025digitaltwin}.

\subsubsection{Distributed actuation}
\label{sec:muscle-model}

Following the octopus arm histology~\cite{Margheri2012histology,tekinalp2024topology}, each arm contains four longitudinal muscles $\LM[1:4]$ for directional bending, one transverse muscle $\TM$ for elongation, and two opposing oblique muscles $\OM[\pm]$ for twisting ~(\Cref{fig2:setup}e--f).
Two groups of ventral suckers, proximal $\Sucker[p]$ and mid-arm $\Sucker[m]$, provide intermittent substrate anchoring.
These nine actuation elements form the actuation set $\mathcal{A}=\{\LM[1],\LM[2],\LM[3],\LM[4],\TM,\OM[+],\OM[-],\Sucker[p],\Sucker[m]\}$.
Each element's activation $\nu^{a}(t)\in[0,1]$ follows a temporal Gaussian pulse specified by its amplitude, width, and peak phase within the locomotion cycle $T$.
Stacking these parameters gives the control $u^{(m)}\in\armctrlspace=[-1,1]^U$ for arm $m$, with dimension $U=\abs{\mathcal{A}}\times3=27$.

For an actuation element $a\in\mathcal{A}$, let $\musclelength[a](s,t)>0$, $\muscletangent[a](s,t)\in\mathbb{S}^2$, and $\musclepositions[a](s,t)\in\reals^3$ denote its local relative length, unit tangent, and cross-sectional offset, respectively.
Omitting the arguments $(s,t)$ for brevity, the contractile force and corresponding force and couple resultants are
\begin{equation}
    \muscleforce[a] = \nu^{a}\hillsmodel(\musclelength[a]), \qquad
    \muscleforces[a] = \muscleforce[a]\muscletangent[a], \qquad
    \musclecouples[a] = \musclepositions[a]\times\muscleforces[a],
    \label{eq:main_muscle_model}
\end{equation}
where $\hillsmodel:\reals_{+}\to\reals_{+}$ is the force-length response (Hill's model) at full activation.
Adding the muscle contributions to the arm's passive elastic response gives the total internal force and couple resultants,
\begin{align*}
    \internalforces = \internalforces^{\text{p}} + \sum_{a\in\mathcal{A}}\muscleforces[a],\quad\internalcouples = \internalcouples^{\text{p}} + \sum_{a\in\mathcal{A}}\musclecouples[a],
\end{align*}
where $\internalforces^{\text{p}}$ and $\internalcouples^{\text{p}}$ are the passive elastic resultants.
These internal loads couple muscle actuation~\eqref{eq:main_muscle_model} to arm deformation through the Cosserat dynamics~\eqref{eq:main_cosserat_dynamics}.
Together with substrate contact, they produce the displacement $\disps(u)$ and rotation $\dispPhi(u)$ used to evaluate control $u$ in objective function \eqref{eq:objective_function}.





\subsubsection{Baselines and metrics}
\label{sec:baselines-and-metrics}

Experiments first assess full-direction control performance after symmetry folding and interpolation, together with path-tracking on a sequence of waypoints. Second, we analyze optimization progress against similar baselines: discovery of high-objective controls and maintaining variability in the generated controls.

We compare DUO with DiffBBO and DiBO, the closest iterative diffusion-optimization methods. Multi-CMA-ES~\cite{Hansen2016TheCE} is included as an intuitive extension that obtains multiple solutions from a population-based optimization.
Lastly, we compare with AutoQD~\cite{hedayatian2026autoqd}, a variant of a quality--diversity method that automatically constructs the behavior descriptors, applicable to our problem.
Each baseline is implemented from its manuscript and adapted to our environment setup. 
For methods using neural networks, we match the architecture and parameter count.
All optimization uses the same simulator-evaluation budget.

\subsubsection{Adaptation evaluation}
\label{sec:adaptation_setup}


Lastly, we demonstrate an essential benefit of adopting a diffusion-based method: adaptation under actuation constraints, without retraining.
Using an editing technique similar to \emph{inpainting}~\cite{lugmayr2022repaint,meng2022SDEdit},
additional constraints can be imposed on a previously generated control by iterating denoise-constrain-noise steps~\cite{lugmayr2022repaint} ~(\Cref{fig3:method}c).

For evaluation, we selected three scenarios that impose a significant, yet recoverable, impact on the controller.
First, a $25\%$ reduction in all muscles, simulating a mild, uniformly degrading case that would require a small amount of control reconfiguration to recover.
Next, a $75\%$ reduction in suction on all arms, mimicking ineffective suction, as when a surface affords only a weak grip such as a simple granular surface.
Lastly, complete suppression of actuation on four alternating arms ($m=1,3,5,7$) to simulate partial arm availability.


\begin{figure*}[t!]
    \vspace{5pt}
    \centering
    \includegraphics[width=\linewidth]{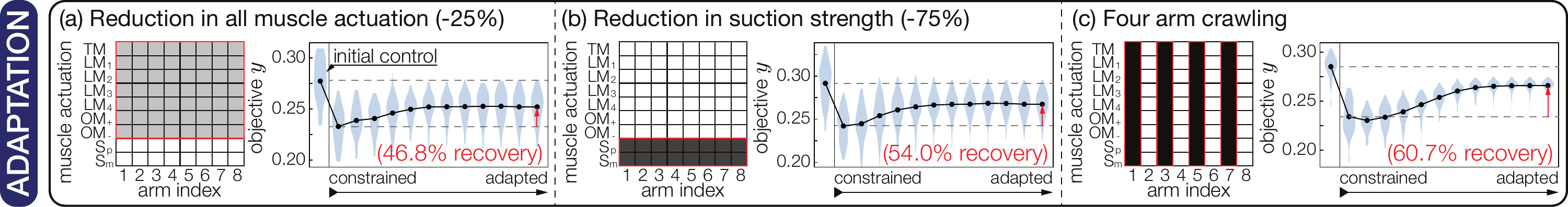}
    \caption{\textbf{Adaptive control editing under actuation constraints.}
    Performance objective recovery outcomes produced by the trained and fixed DUO denoiser:
    (a) $25\%$ muscle-amplitude reduction across all arms;
    (b) $75\%$ suction-amplitude reduction across all arms;
    (c) zero muscle and suction amplitudes on arms $1$, $3$, $5$, and $7$.
    On the left, dark cells mark constrained channels.
    On the right, blue violin plots show retained-outcome objective distributions over $64$ generated controls; black markers show means. 
    The plotted points illustrate the performance across the adaptation process: starting from the initial control, applying the constraint, and showing the subsequent edited versions across the iterative denoising steps.
    Red percentages report the mean of recovery, $(y_{\mathrm{adapted}}-y_{\mathrm{constrained}})/(y_{\mathrm{initial}}-y_{\mathrm{constrained}})\times100\%$.
    }
    \label{fig5:result2}
    \vspace{-17pt}
\end{figure*}


\subsubsection{Hyperparameters}
\label{sec:hyperparameters}

All evaluations run for $100$ iterations of $\querysize=1{,}800$ evaluations.
For DUO, we use $\ensemblesize=J=8$ ensemble sizes and generate candidates at a ratio $\nicefrac{\widehat{\querysize}}{\querysize}=3$, and sample target objective candidates $\{y^i\}_{i=1}^{\widehat{\querysize}}$ uniformly from $\fitnessset_{\optiter}=[0.6y_{\max,\ell},y_{\max,\ell}]$ for each iteration $\ell$.
After a hyperparameter sweep, we use $\lambda=0.5$, $\kappa=1$, and $\eta=0.25$, which attained the highest performance while maintaining variation among the controls explored during search.


We train each denoiser for $10{,}000$ gradient steps per iteration, using
$K=64$ diffusion steps with a linear variance schedule~\cite{ho2020denoising},
conditioning dropout is $0.15$, and classifier-free guidance scale is $2$~\cite{ho2021classifierfree}.
Remaining hyperparameter details are provided in the code repository.

\section{RESULTS AND DISCUSSION}
\label{sec:results-discussion}


\subsubsection{Full-direction crawling}
\label{sec:results_symmetry_control}



\Cref{fig4:result1}a--b evaluates the controller over the canonical forward sector.
Without the boundary objective, interpolation severely degrades in the interior of the sector, in both the crawling objective and the penalty terms.
This deviation is more pronounced in waypoint tracking~(\Cref{fig4:result1}c).
Adding the boundary objective, on the other hand, loses a little performance at the center, but yields more uniform crawling in all directions, with reduced angular deviation and body rotation~(\Cref{fig4:result1}b).
Together, these results support including the boundary objective to improve directional consistency between arm-aligned commands.

\Cref{fig4:result1}d illustrates crawling along a curved target path, with successive body configurations showing coordinated arm motion as the commanded direction changes.
All trajectories from $128$ generated controls show successful tracking and reach the end. This example demonstrates how the directional controller construction supports successive changes in crawling direction.

\subsubsection{Multimodal control discovery}
\label{sec:results_coordination}






The actuation profiles in \Cref{fig4:result1}e, averaged across the arms, provide a muscle-coordination view of the discovered controller diversity.
The control families follow distinct group across all activation types, rather than differing only by small deviations from a shared profile.
Interestingly, LM and OM actuation level remain relatively uniform across controls, while more distinct characteristics emerge in the utilization of TM and suction.
Together, these results indicate that the learned distribution contains multiple coordination patterns for the crawling task.
They do not, by themselves, establish mechanically discrete gaits, but instead reveal distinct control and recruitment patterns within a common locomotion task.

The progression of the training is visualized through  training curve and a principal-component embedding, with color indicating the achieved objective ~(\Cref{fig4:result1}f--g).
Initially at $\ell=0$, sampled controls occupy a compact, low-objective region.
By $\ell=30$, the objective has risen rapidly and the distribution has branched into several clusters.
As training proceeds, both the PC1--PC2 projection and variation in the training curve indicate continued exploration: generated controls continue to spread outwards while previously discovered solutions are retained and exploited.
This spread persists through the end of training $(\ell=100)$ rather than collapsing onto the single best control, with some generated controls remaining in slightly suboptimal regions.
Thus, DUO retains high-performing controls while continuing to query nearby, non-maximal regions during online optimization.


\subsubsection{Comparison with other methods}
\label{sec:results_method}
Training progress and comparisons with related methods are reported in \Cref{fig4:result1}f. The plot shows the min--max range of the objective together with the corresponding batch maximum.

DiBO and DUO reach the highest values, with DiBO converging slightly earlier.
DUO continues to evaluate controls over a wider objective range than DiBO.
This range implies that the acquisition procedure does not concentrate too quickly, and that evaluation includes a wider range away from the optimum.
Once the maximum is reached, the variance increases slowly as the search continues actively in under-explored regions.

DiffBBO improves more gradually and reaches a lower batch maximum of approximately $0.27$, and exhibits a smaller min-max range compared to DUO.
Additionally, while Multi-CMA-ES demonstrates high diversity, its batch objective remains below that of the diffusion-based methods under the matched evaluation budget.

AutoQD improves more slowly, attaining a retained best-so-far of $0.173$ later than the diffusion-based methods, but it searches more thoroughly.
Its quality--diversity mechanism ranks candidates by improvement within cells of a learned behavior archive rather than by raw objective alone~\cite{hedayatian2026autoqd}, so emitters continue to probe underrepresented regions---the corners of the archive---after a high-performing control has been found.
Periodic updates of the behavior descriptor remap the archive and restart the emitters, introducing discontinuities between proposal batches; the batch mean and maximum can therefore decline while the retained best-so-far is unchanged.
That late decline reflects continued coverage of the behavior space and the quality--diversity trade-off, not loss of the archive's best controller.
Although AutoQD's batch maximum remains below DUO, the curves indicate broader coverage of AutoQD's archive.

\begin{table}[b]
\vspace{-20pt}
\centering
\caption{Ablation study of DUO components.
Values are mean over five independent runs, each of $100$ iterations.}
\label{tab:DUO_ablation}
\small
\setlength{\tabcolsep}{4pt}
\begin{tabular}{lcc}
\toprule
Variation
& Final best
& $\Delta$ (\%) \\
\midrule
\textbf{DUO} & $\mathbf{0.384}$ & -- \\
- sample $u^i$ from $p_{\theta_e}(\cdot|y_\text{max})$ instead & $0.256 $ & $-33.3$ \\
- without $\alpha^\text{control}$ & $0.243 $ & $-36.7$ \\
- without $\alpha^\text{target}$ & $0.328$ & $-14.6$ \\
\bottomrule
\end{tabular}
\end{table}

\subsubsection{Adaptation}
\label{sec:results_adaptation}

\Cref{fig5:result2} reports adaptive control editing under the three different actuation constraints, with the trained denoiser held frozen.
The violin plots show how generated controls perform through editing with positive recovery.
In all three cases, editing restores about half or more of the lost objective, including when four alternating arms are unavailable.
These results support constrained adaptability in zero-shot: the learned distribution contains compatibility with the new limits, therefore enabling recovery without additional retraining.
Further, recovery under constraints never imposed during training---especially the complete inactivation of four arms---shows that the learned denoiser can reconfigure its control strategy, a mechanism distinct from simply retrieving a stored solution from the evaluation data.



\subsubsection{Ablation study}
\label{sec:results_ablation}

Three ablation studies isolate the importance of each mechanism introduced in DUO~(\Cref{tab:DUO_ablation}).
First, we compare against using a single conditioning objective, set to the maximum value in the observed dataset, as commonly used in optimization without a diversity focus or in offline settings.
Conditioning exclusively on the maximum is reasonable when seeking a single high-performing control, but limits exploration and can prematurely concentrate the search around local optima.
Indeed, this replacement reduces the final objective by $33.3\%$, indicating that conditioning over a broader objective range promotes effective exploration and better coverage across the $y$-conditioning space.

We next isolate the two acquisition mechanisms that distinguish DUO from a single-path diffusion query.
Removing the term $\acquisition^{\mathrm{control}}$ produces a $36.7\%$ reduction, while removing the term $\acquisition^{\mathrm{target}}$ reduces the objective by $14.6\%$.
Together, the three ablations show complementary roles: the ranged target-member pool maintains search coverage, control acquisition directs the simulation budget toward promising controls, and target acquisition keeps queries within reliable objective levels.
These mechanisms balance exploration with selective evaluation, providing informative data throughout online training.





\section{CONCLUSION}
\label{sec:conclusion}

In soft robotics, control redundancy gives rise to a wide variety of solutions, which diffusion models offer substantial potential to handle.
In support of this, we present Diffusion-based Uncertainty-aware Optimization (DUO), an efficient online data-collection procedure for training a denoiser.
This method learns to generate high-performing controls without relying on prior demonstrations.
Together with radial symmetry reduction, we obtain diverse muscle-arm coordination strategies for planar crawling in all commanded directions for the CyberOctopus.
Further, diffusion-based control editing enables partial performance recovery under previously unseen actuation constraints without retraining.
These results demonstrate how DUO turn the representational capacity of diffusion models into a practical resource for both control and adaptation.

More broadly, these results suggest the potential of diffusion-based controller for interactive soft-robot operation, beyond isolated experimental settings, and toward open-environment interaction.
Future work will validate the generated controllers and adaptation strategies on physical hardware~\cite{nazeer2026benchmarking}, incorporate sensory feedback~\cite{wang2025neural} for more complex tasks, and extend the approach to three-dimensional locomotion~\cite{wu2024octopus}.







\bibliographystyle{IEEEtran}
\bibliography{bib/SHK,bib/KK,bib/misc,bib/HSC}  

\setcounter{section}{0}
\renewcommand{\thesection}{\Alph{section}}


\end{document}